\documentclass[10pt,twocolumn,letterpaper]{article}

\usepackage{cvpr} 

\usepackage{times}
\usepackage{epsfig}
\usepackage{graphicx}
\usepackage{amsmath}
\usepackage{amssymb}
\usepackage{multirow}
\usepackage{comment}
\usepackage{boldline,multirow}
\usepackage{amssymb}
\usepackage{pifont}
\DeclareMathOperator*{\argmax}{\arg\max}
\newcommand{\cmark}{\ding{51}}%
\newcommand{\xmark}{\ding{55}}%
\usepackage{comment}

\usepackage{xcolor}

\usepackage[pagebackref=true,breaklinks=true,colorlinks,bookmarks=false]{hyperref}
\usepackage[capitalize]{cleveref}
\crefname{section}{Sec.}{Secs.}
\Crefname{section}{Section}{Sections}
\Crefname{table}{Table}{Tables}
\crefname{table}{Tab.}{Tabs.}

\newcommand\codelink{https://github.com/YihongSun/Bayesian-Amodal}

\def\cvprPaperID{3504} 
\def\confName{CVPR}
\def\confYear{2022}

\begin{document}

\title{Amodal Segmentation through Out-of-Task and\\ Out-of-Distribution Generalization with a Bayesian Model}  

\author{Yihong Sun\;\; Adam Kortylewski\;\; Alan Yuille \\
\\
Johns Hopkins University\\
}

\maketitle

\begin{abstract}
Amodal completion is a visual task that humans perform easily but which is difficult for computer vision algorithms.
The aim is to segment those object boundaries which are occluded and hence invisible. 
This task is particularly challenging for deep neural networks because data is difficult to obtain and annotate.
Therefore, we formulate amodal segmentation as an out-of-task and out-of-distribution generalization problem. 
Specifically, we replace the fully connected classifier in neural networks with a Bayesian generative model of the neural network features. The model is trained from non-occluded images using bounding box annotations and class labels only, but is applied to generalize out-of-task to object segmentation and to generalize out-of-distribution to segment occluded objects.
We demonstrate how such Bayesian models can naturally generalize beyond the training task labels when they learn a prior that models the object's background context and shape.
Moreover, by leveraging an outlier process, Bayesian models can further generalize out-of-distribution to segment partially occluded objects and to predict their amodal object boundaries.
Our algorithm outperforms alternative methods that use the same supervision by a large margin, and even outperforms methods where annotated amodal segmentations are used during training, when the amount of occlusion is large. Code is publicly available at \href{\codelink}{\codelink}.
\end{abstract}

\begin{figure}
    \centering
    \includegraphics[height=5.82cm]{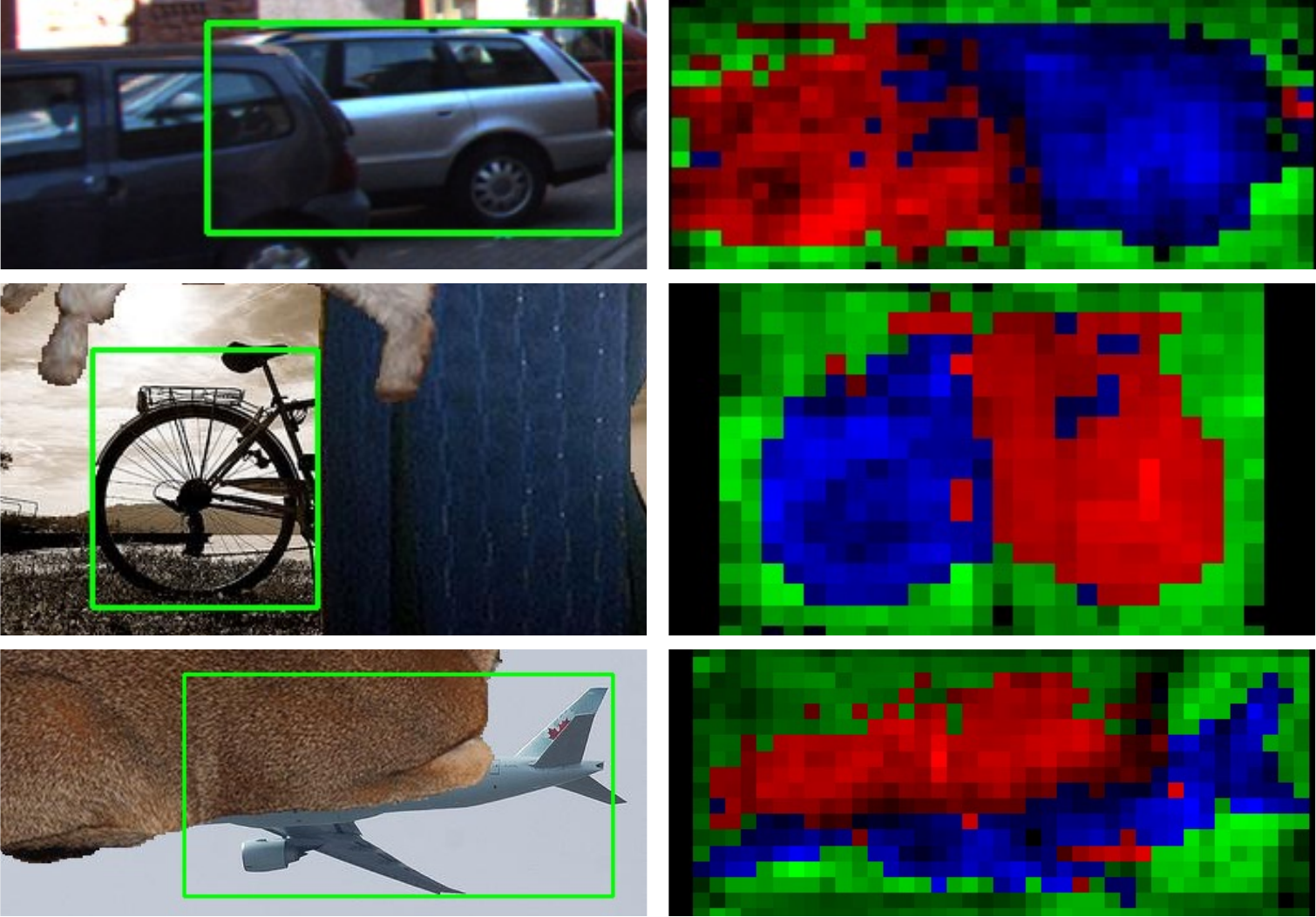}
    \caption{Our Bayesian model takes the object bounding box as input and estimates the three segmentation masks on the right: the visible object parts in blue, the invisible object parts in red, and the background context in green. The model is fully probabilistic, and the pixel brightness shows the confidence of the model prediction.} 
    \label{fig:teaser}
\end{figure}

\section{Introduction}
In our everyday life, we often observe partially occluded objects. Humans can reliably recognize the visible parts of an object and use them as cues to estimate the occluded parts. 
This perception of the object's complete structure under occlusion is referred to as \textit{amodal perception} \cite{amodal_cog}. 

In computer vision, amodal segmentation is important to study, both for its theoretical values and real-world applications. 
The main limitation of current approaches is the requirement of detailed supervision of amodal object masks either through human annotation \cite{li2016amodal,kins, follmann2019learning} or by generating artificially occluded images \cite{deocclusion_2020_CVPR}.
Moreover, these methods assume that the object class of the occluder is known at training time. This is an important limitation in real-world applications, such as autonomous driving, where potential occluders can be any kind of real-world object.

We formulate amodal segmentation as an out-of-task and out-of-distribution generalization problem, where a Bayesian generative model is trained from non-occluded objects with bounding boxes and class annotations only, but generalizes to amodal segmentation of partially occluded objects (Figure \ref{fig:teaser}).
Intuitively, our model can be understood as a convolutional neural network, in which the fully-connected classification head is replaced with a Bayesian generative model of the neural features. 
During inference, the latent model parameters (i.e. object class and amodal segmentation) are estimated such that the features of the input image are explained by the Bayesian model with maximum likelihood. 
The invariance properties of the neural features enable us to avoid explicitly modeling nuisances such as small deformations or illumination changes. 

Our work builds on recent approaches of learning generative models of neural network features for image classification \cite{compnet_cls_cvpr20,compnet2020journal}, and extends these in several ways to enable amodal segmentation.
In particular, we extend the network architecture with a generative model of the object's background context, as well as a prior of the object shape. 
Unlike standard Deep Network approaches, this makes the notion of the background context and the object shape explicit.
Together, these priors enable our model to be trained from bounding box and class supervision only and generalize out-of-task to object segmentation.
The Bayesian model is combined with an outlier process to make it robust to partial occlusion. 
The outlier process enables us to formulate amodal segmentation as an out-of-distribution task, where the model is trained from non-occluded images, but generalizes to images with partially occluded objects. 
We discuss how the full Bayesian model can be learned using maximum likelihood estimation with an EM-type algorithm.
We also demonstrate that a joint end-to-end fine-tuning of the Bayesian model and the convolutional feature extractor further improves the performance by a steady margin.

Our experiments on all common datasets for amodal segmentation, KITTI Instance dataset (KINS) \cite{kins}, COCO Amodal \textit{cls.} \cite{cocoacls} and Occluded-PASCAL3D+ \cite{compnet_det_cvpr20}, show that our Bayesian approach outperforms related weakly-supervised work by a large margin and even outperforms fully supervised methods when the amount of occlusion is large. 
In summary, we make several contributions:

\begin{enumerate}
\item We formulate \textbf{amodal instance segmentation as an out-of-task and out-of-distribution generalization problem} with a Bayesian generative model.
\item Our Bayesian model is learned from bounding box and class labels only and \textbf{outperforms alternative weakly-supervised methods by a large margin} and even outperforms supervised methods (where annotated amodal segmentations are used during training) when the amount of occlusion is large. 
\item To the best of our knowledge, our model is the first for amodal segmentation that \textbf{generalizes to previously unseen occluders}.
\end{enumerate}

\section{Related Work}

\textbf{Amodal Segmentation.} 
One of the first works in amodal segmentation is proposed by Li \textit{et al.} \cite{li2016amodal} with an artificially generated occlusion dataset. Recently, the KINS \cite{kins} and Amodal COCO \cite{amodal_coco} datasets were introduced, which contain real-world occlusion and human estimated amodal segmentation masks.
Related work on amodal segmentation follow a fully supervised approach, where either human-estimated amodal segmentation annotations are used \cite{kins, follmann2019learning, review-d}, or synthetic occlusions are created to create training data \cite{deocclusion_2020_CVPR, review-a, review-b}. 
However, these approaches make implicit assumptions about the amount of occlusion at test time, or even require the class of the occluder to be known at test time \cite{deocclusion_2020_CVPR, review-a}. 
In contrast, we introduce a Bayesian approach to this problem which is trained from non-occluded objects only and does not require any amodal supervision.

\textbf{Robustness to Occlusion with Bayesian Models.} 
Amodal segmentation is a relatively new research direction, but research on robustness to partial occlusion has received a lot of attention. In the following we focus solely on works that directly relate to ours.
Recent studies \cite{hongru,kortylewski2019compositional} showed that typical deep learning approaches to image classification are significantly less robust to partial occlusions than human vision. 
In contrast, Bayesian approaches are significantly more robust to partial occlusion, as shown in the domains of image classification \cite{compnet_cls_cvpr20}, pose estimation \cite{wang2021nemo}, general object detection \cite{compnet_det_cvpr20, compnet2020journal}, scene understanding \cite{abs_scene, abs_scene2}, face reconstruction \cite{egger2018occlusion} and human detection \cite{girshick2011object}.  
In this work, we generalize such Bayesian generative models of neural features to amodal segmentation by leveraging estimated per-pixel occlusion statistics. 
Notably, our work is related to Bayesian generative approaches that were developed in the pre-deep-learning-era \cite{trad_work}. However, our combination of modern deep learning with Bayesian generative models, enables us to generalize to very complex data with weak supervision only.
%

\textbf{Weakly-supervised Segmentation.} Due to the demanding task of acquiring expensive per-pixel annotations, many weakly-supervised instance segmentation methods have emerged that leverage cheaper labels, including image-level \cite{ahn2019weakly, cholakkal2019object, arun2020weakly, laradji2019masks, zhou2018weakly, zhu2019learning} and box-level annotations \cite{khoreva2017simple, bbtp_hsu2019weakly}.
Notably, Zhou \textit{et al.} \cite{zhou2018weakly} propose to use image-level annotations to supervise instance segmentation by exploiting class peak responses to enable a classification network for instance mask extraction.
Additionally, Hsu \textit{et al.} \cite{bbtp_hsu2019weakly} uses box-level annotations to achieve instance segmentation by exploiting the bounding box tightness prior.
Finally, Shapemask \cite{kuo2019shapemask} addresses instance segmentation of objects with novel categories without mask annotations. Through exploiting the shape priors of known objects learned from ground truth masks, Shapemask learns the object shape and generalizes instance segmentation to novel categories. In contrast, our proposed model is able to learn shape priors without any pixel-level supervision.

\section{A Bayesian Model for Amodal Segmentation}
\label{sec:compnet}

In the following, we introduce our model by first describing its input, then a simplified version of the model, and finally we proceed to develop our full Bayesian model. The structure of this section also serves to clarify the similarity and differences between our model and related work on generative models of neural network features. 
\subsection{The input to our model: Neural Features}
\label{sec:input} 
Our model takes as input a feature map $\bar{F}= \psi(I,\zeta)$ at the top convolution layer of a Deep Neural Network where $I$ is the input image and $\zeta$ are the weights of the convolutional layers. The network weights can be learnt by pre-training on ImageNet or can be directly trained end-to-end. The key property of these feature vectors is that they tend to be invariant to unimportant details of the object, which makes it easier to learn a Bayesian generative model compared to using RGB pixels as input. We denote the features within a given bounding box $\mathcal{D}$ by $F = \{f_a: a \in \mathcal{D}\}$, where $a$ denotes the position on the lattice within the bounding box. Hence, $F$ denotes a cropped subset of the feature map $\bar{F}$.

\subsection{A Simplified Generative Model}
\label{sec:simple}

We now discuss a simplified Bayesian generative model of the feature vectors and discuss how it can be modified to make it robust to occluders and how it can be learned. For each object, we assume that the features are generated by a mixture of distributions, which roughly correspond to the viewpoint of the objects (Equation \ref{eq:1}). This is similar to deformable part models \cite{felzenszwalb2005pictorial,felzenszwalb2010cascade} where these mixtures also have to be learnt without supervision. However, these approaches are not generative and do not address the problem of partial occlusion.


The simplest generative probability model, which corresponds to the model introduced in related work \cite{compnet_cls_cvpr20,compnet2020journal}, specifies a probability distribution:
\begin{align} 
&P(F|y) \hspace{-0.04cm} = \hspace{-0.08cm} \sum _m p(F|y,m) P(m) \hspace{-0.04cm} = \hspace{-0.08cm} \sum_m \hspace{-0.06cm} \prod_{a \in \mathcal{D}} \hspace{-0.07cm} P_a(f_a|y,m) P(m)\label{eq:1},\\
&P_a(f_a|y,m)\hspace{-0.04cm}=\hspace{-0.04cm}P_a(f_a|\mathcal{A},\Lambda) \hspace{-0.04cm}=\hspace{-0.04cm}\sum_k \alpha_{i,k}^{y,m} P(f_a|\sigma_k, \mu _k)\label{eq:2},\\
&p(f|\sigma_k, \mu _k) = \frac{e^{\sigma_k \mu_k^T f}}{Z(\sigma_k)}, ||f|| = 1, ||\mu_k|| = 1\label{eq:3},
\end{align}
\noindent 
where $y$ denotes the object class and $m$ refers to the mixture component. The number of mixtures is fixed a-priori and the mixture components are learnt in an unsupervised manner (similar as in Gaussian mixture models). $P(m)$ is an uniform prior over the mixture components, $\mathcal{A}= \{\alpha_{i,k}^{y,m}\}$ are mixture coefficients and $\Lambda = \{\sigma _k, \mu _k\}$ are the parameters of a von-Mises-Fisher (vMF) distribution (Equation \ref{eq:3}). We choose a vMF distribution, because normalizing the feature vectors to unit norm makes it more feasible to estimate the model parameters in the high dimensional feature space of neural networks (note that the dimensionality of a feature vectors $f_a$ in higher convolutional layers is typically 1024).

\textbf{Learning the model parameters.}
In most of this paper, we assume that the parameters $\zeta$ of the Deep Network have been learnt in advance. This enables the remaining parameters of the model to be learnt by standard Bayesian methods using Maximum Likelihood via the Expectation-Maximization (EM) algorithm. 
Since our Bayesian model is fully differentiable, we will also discuss an alternative end-to-end learning method which learns all model parameters jointly in Section \ref{sec:e2e}. The end-to-end training improves over the ML solution by a small but steady margin. 

As shown in \cite{compnet_cls_cvpr20,compnet2020journal}, the parameters $\Lambda$ correspond intuitively to a vocabulary of parts of the objects and can be learnt simply by the EM algorithm \cite{dempster1977maximum} initialized by the K-Means++ clustering algorithm \cite{arthur2006k}. The probability distributions $P(F|y)$ can be learnt using maximum likelihood to estimate the parameters $\mathcal{A}$. This also only requires the simple application of the EM algorithm because of the latent mixture variables $m$. For the sake of clarity, we refer the reader to our implementation for details on the EM learning, as the application of EM algorithm is a standard process for statistical distributions with unobserved latent variables. 

Finally, the inference process is a feed-forward pass through the network to estimate ${\hat y} = \argmax_y P(F|y)$. 

\textbf{Occlusion modeling.} To make this model robust to occluders and enable it to generalize out-of-distribution when trained with non-occluded objects, the generative model is modified by adding an outlier process to take the form:
\begin{equation} 
\label{eq:occ}
P(F|y) = \sum_m \prod _{a \in \mathcal{D}} P_a(f_a|y,m) ^{z_a}Q(f_a)^{1 - z_a}P(m)P(\vec z),
\end{equation}
\noindent where $Q(f_a)$ is a von Mises Fisher distribution for a feature generated by an occluder estimated from unannotated images \cite{kortylewski2017model,egger2018occlusion}. The latent variable $z_a \in \{0,1\}$ indicates whether pixel $a$ is visible or occluded ($z_a=1, 0$ respectively) and the prior $P(\vec z)$ indicates the prior probability of a pixel being visible. This enables the model to not only be robust to occluders but to also simultaneously estimate the locations of the occluders $\{a \in \mathcal{D}: z_a=0\}$ \cite{compnet_cls_cvpr20,compnet2020journal}, in addition to the object $y$, the mixture component $m$. 

\begin{figure}
    \centering
    \includegraphics[height=1.735cm]{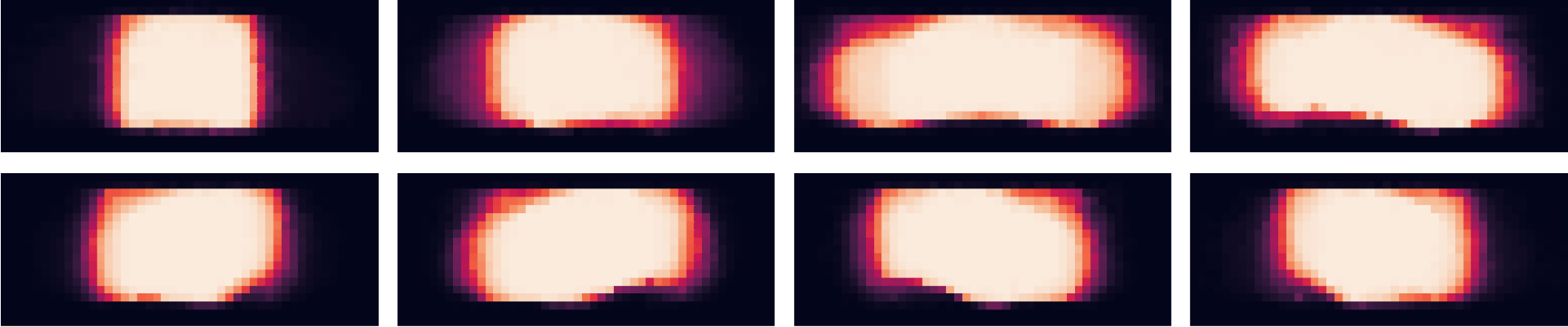}
    \caption{Compositional Shape Priors $P(\vec w|y,m)$. $M=8$ compositional shape priors belonging to the car class are shown. Note that in every prior, shape and 3D pose encoding are learned by leveraging bounding box supervision only.}
    \label{fig:3d_prior}
\end{figure}

\subsection{A Bayesian Model for Amodal Segmentation}
\label{sec:ourmodel}

A limitation of the simplified model described in the previous section is that it cannot segment the object, because it does not separate the foreground region corresponding to the object and a background region corresponding to the local background context of the object (e.g., the background context of an airplane will typically be sky). 
This motivates us to extend the model by introducing new latent variables $\{w_a\}$ to indicate foreground/background which are learnt without additional supervision. 
We start by extending the generative model introduced in Equation \ref{eq:1} to be of form:
\begin{align}
P_a(f_a|y,m,w_a) = &P_a(f_a|y,m) ^{w_a} B_a(f_a|y,m) ^{1-w_a}\label{eq:fgbg}\\
&\times P_a(w_a|y,m) \nonumber
\end{align} 
\noindent where $w_a \in \{0,1\}$ is a latent variable indicating whether the pixel is foreground or background context ($w_a =1,0$ respectively). Here $\{P_a(f_a|y,m)\}$ and $\{B_a(f_a|y,m)\}$ are models for the foreground and background pixels respectively. They are specified by the foreground and background mixtures of von Mises Fisher distributions, respectively, with the same form as in Equation \ref{eq:2}. 

\textbf{Shape Modeling.} We introduce shape priors $P(\vec w|y,m) = \prod _{a \in \mathcal{D}} P_a(w_a|y,m)$, a learned 2D spatial map conditioned on the object category $y$ and the class mixture $m$ for the foreground/background masks as shown in Figure \ref{fig:3d_prior}.
Intuitively, they model the expected object shape for each mixture model $m$, and will enable the model to predict the object shape behind an occluder, as discussed in the next section. The structure of the shape priors in Figure \ref{fig:3d_prior} shows that the mixture components $m$ approximately represent different 3D object poses. 
Finally, this gives a generative model of the data:
\begin{equation} 
P(F|y) = \sum _{m, \vec w} P(F|y,m,\vec w) P(m)P(\vec w|y,m),
\end{equation}
The model can be learned by maximizing the log-likelihood of the training data with respect to $\Lambda, \mathcal{A}, P(\vec w|y,m)$. This requires using the EM algorithm since the model contains latent variables for the mixtures $m$ and the foreground/background variables $\{w_a\}$. During learning we use the standard maximum likelihood measures for vMF distributions \cite{banerjee2005clustering}, and initialize EM for the class mixtures using spectral clustering, as in \cite{compnet_cls_cvpr20,compnet2020journal}. To initialize the foreground/background variables $\{w_a\}$, we first initialize the background distribution $B_a$ from unannotated data (similar to estimating the distribution for the occluder $Q$) and initially assume that everything is foreground (i.e. $w_a =1 \forall a$). 


\textbf{Occlusion Modeling.} To extend this model to deal with occlusion, we also introduce an outlier process. As described in Equation \ref{eq:occ}, we introduce binary latent variables $\{z_a\}$ where $z_a \in \{0,1\}$ indicates whether pixel $a$ is visible ($z_a=1$) or occluded ($z_a=0$). We introduce an occluder distribution $Q(.)$ which is a von Mises Fisher distributions $Q(f_a) = \frac{e^{\sigma \mu^T f_a}}{Z(\sigma)}, ||f_a|| = 1, ||\mu|| = 1$ whose parameters are learnt from features in unannotated images.  We specify, but do not learn, a  prior $P(\vec z) = \prod _{a \in \mathcal{D}}P(z_a)$ where $P(z=0)$ is a rough measure of how much occlusion we want the algorithm to be able to deal with. 

\textbf{Displacement Modeling.} We also introduce a displacement variable $c$ that models the displacement between the center of the bounding box and the center of the object. This is necessary because for partially occluded objects the bounding box only covers the visible part of the object, but amodal segmentation requires the model to predict the invisible object boundary. This gives a model of form:

\begin{align} 
P(F|y) = \sum_m \prod _{a \in \mathcal{D}} P_{a-c}(f_a|y,m) ^{w_a z_a} \label{eq:center}\\ 
\times B_{a-c}(f_a|y,m)^{(1-w_a)z_a} Q(f_a) ^{(1-z_a)}. \nonumber \end{align} 
\begin{equation} 
P(\vec w|y,m,c) = \prod _{a \in \mathcal{D}} P_{a-c}(w_a|y,m)
\end{equation}

Using this model, we can estimate the optimal object class $y$, class mixture $m$, object center $c$, occlusion map $\{a \in \mathcal{D}: z_a=0\}$, and foreground map $\{a \in \mathcal{D}: w_a=1\}$. 
This inference process can be implemented efficiently as a feed-forward neural network and we provide a publicly available implementation  \footnote{\href{\codelink}{\codelink}}.


\subsection{Amodal Segmentation with Our Model}
\label{sec:inference}
\begin{figure}
    \centering
    \includegraphics[width=\linewidth]{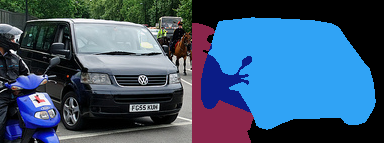}
    \caption{Illustration of the four states that a pixel can be in: \textit{Visible Foreground} ($w_a=1, z_a=1$) in light blue, \textit{Occluded Foreground} ($w_a=1, z_a=0$) in dark blue, \textit{Visible Background}  ($w_a=0, z_a=1$) in black, and \textit{Occluded Background} ($w_a=0, z_a=0$) in red. Consequently, amodal segmentation of an occluded object is thus defined as \textit{Visible Foreground} $\cup$ \textit{Occluded Foreground}. }
    \label{fig:states}
\end{figure}

After estimating distributions for the latent variables $w_a$ and $z_a$, the states of $w_a$ and $z_a$ categorize each image pixel into one of four potential states (Figure \ref{fig:states}). Thus, we can determine the amodal object segmentation by finding the visible and occluded foreground regions as follows.

To estimate the foreground-background segmentation $w_a$, we compute the posterior odds between the foreground and the background probabilities:
\begin{align}
\label{eq:seg_new}
    \hat{w}_a &= 
    \begin{cases}
        1,& \text{if } \frac{P_{a-c}(w_a|y,m) \hspace{0.1cm} P_{a-c}(f_a|y,m)}{(1-P_{a-c}(w_a|y,m)) \hspace{0.1cm} B_{a-c}(f_a|y,m)} > 1\\
        0,              & \text{otherwise.}
    \end{cases}
\end{align}
Similarly, to infer the state of the the occlusion variables $z_a$, we compute the posterior odds between the occlusion and the respective foreground-background probabilities:
\begin{align}
    \hat{z}_a &= 
    \begin{cases}
        1,& \text{if } \frac{p(z_a) P_{a-c}(f_a|y,m) ^{\hat{w}_a} B_{a-c}(f_a|y,m)^{(1-\hat{w}_a)}
        }{(1-p(z_a)) Q(f_a)} > 1\\
        0,              & \text{otherwise.}
    \end{cases}
\end{align}{}
Together the states of $\hat{z}_a$ and $\hat{w}_a$ allow the model to infer visible instance segmentation $\hat{M}_I=\{a : w_a=1, z_a=1\}$ and amodal segmentation $\hat{M}_A=\{a : w_a=1\}$, as depicted in Figure \ref{fig:states}. Qualitative visualization of this inference process are illustrated in Figure \ref{fig:teaser}, where the relative confidence of the visible foreground $\{a : w_a=1, z_a=1\}$, occluded foreground $\{a : w_a=1, z_a=0\}$, and background $\{a : w_a=0\}$ are represented by 3-color intensities.

\subsection{End-to-End Training}
\label{sec:e2e}
When learning the parameters of the Bayesian model with the EM algorithm, we assume that the parameters of the feature extractor $\zeta$ have been initialized and fixed. 
This is achieved by pre-training the feature extractor for image classification using a fully-connected prediction layer and then replacing it with our Bayesian generative model. 
But our Bayesian model is fully differentiable and hence we can fine-tune the feature extractor and the Bayesian predictor jointly with backpropagation. This enables the feature extractor to adapt to the new predictor, which increases the models performance by a steady margin.

The objective for the end-to-end training includes a cross-entropy classification loss
$\mathcal{L}_{cls}(\hat{y},y)$ using the negative log-probability
$\hat{y}=\argmax_y-\log P(F|y)$, where $\hat{y}$ is the predicted class label and $y$
is the true class labels. 
Following ~\cite{compnet_cls_cvpr20,compnet2020journal}, the parameters of the Bayesian model need to be trained with an additional loss ($\mathcal{L}_{ml}$) such that the Bayesian model retains a maximum likelihood of the data when the feature extractor is updated. Finally, 
we include an additional prior $\mathcal{L}_{seg}(\hat{M}_I,b)$ as proposed by \cite{bbtp_hsu2019weakly} which encourages label consistency within neighboring pixels of the estimated segmentation mask. 
We train all parameters of our model end-to-end with $\gamma_1$ and $\gamma_2$ controlling the trade-off of the loss terms:
\begin{align}
    \label{eq:loss}
    \mathcal{L} \hspace{-.055cm}=\hspace{-.055cm} \mathcal{L}_{cls}(\hat{y},y)\hspace{-.055cm} + \hspace{-.055cm} \gamma_1 \mathcal{L}_{ml}(\Lambda, \mathcal{A},{\vec w}) \hspace{-.055cm} + \hspace{-.055cm} \gamma_2 \mathcal{L}_{seg}(\hat{M}_I,b) 
\end{align}
We note that our end-to-end trained model retains the ability to generalize out-of-distribution to partially occluded objects without ever observing partial occlusion during training. This is in contrast to standard deep networks, which do not generalize in OOD scenarios. The reason is that our model remains a generative model that is optimizing a Maximum Likelihood objective, and hence can become robust to occlusion, when equipped with an outlier process.
\section{Experiments}
\label{sec:exp}
We evaluate amodal segmentation performance of our model against a segmentation mask-supervised and a bounding box-supervised baseline on three popular amodal segmentation datasets. Due to the differences between the two baselines, we conduct experiments under two setups, one where the location of the object center is known and the other where the object center needs to be estimated.
\begin{table*} 
\centering
\tabcolsep=0.18cm
\begin{center}
\begin{tabular}{V{2.5}lV{2.5}cV{2.5}cV{2.5}cV{2.5}c|c|cV{2.5}c|c|cV{2.5}c|c|cV{2.5}cV{2.5}}
		\hline
		\multicolumn{14}{V{2.5}cV{2.5}}{Amodal Segmentation on OccludedVehicles} \\
		\hline	
		\multirow{2}{*}{Methods} 	&\multirow{2}{*}{known $c$} &\multirow{2}{*}{superv.} & FG-0 & \multicolumn{3}{cV{2.5}}{FG-1} & \multicolumn{3}{cV{2.5}}{FG-2} & \multicolumn{3}{cV{2.5}}{FG-3} &\multirow{2}{*}{Mean} \\
	    \cline{4-13}
		 	& & & - & BG-1 & BG-2 & BG-3  & BG-1 & BG-2 & BG-3  & BG-1 & BG-2 & BG-3  & \\    		
		\hline  
		PCNet-M         & \xmark & \textit{mask*} &\textbf{77.6}	&\textbf{70.5}	&\textbf{67.8}	&\textbf{64.9}	&\textbf{65.4}	&\textbf{61.3}	&\textbf{56.9}	&\textbf{59.5}	&\textbf{54.4}	&47.6	&\textbf{62.6}\\
		Ours-ML        	& \xmark & \textit{box} &63.3	&60.2	&59.9	&59.8	&56.9	&55.6	&54.8	&52.6	&50.2	&47.1	&56\\
		Ours-E2E        & \xmark & \textit{box} &63	&59.5	&59.5	&59.5	&56.2	&55.9	&55.6	&51.9	&50.6	&\textbf{48.3}	&56\\
		\hline
		\hline	
		BBTP            & \cmark & \textit{box} &\textbf{66.5}	&\textbf{59.7}	&58.4	&57.9	&54.4	&51	&48.9	&50.4	&44.7	&40.2	&53.2\\
		Ours-ML        	& \cmark & \textit{box} &63.7	&59.4	&59.3	&59.6	&57	&56.6	&56.7	&54.7	&53.5	&53.2	&57.4\\	
		Ours-E2E        & \cmark & \textit{box} &63.9	&\textbf{59.7}	&\textbf{59.6}	&\textbf{59.7}	&\textbf{57.2}	&\textbf{56.8}	&\textbf{56.8}	&\textbf{55}	 &\textbf{53.9}	&\textbf{53.4}	&\textbf{57.6}\\
	    \hline
\end{tabular}
\end{center}

\caption{Amodal Segmentation performance evaluated on OccludedVehicles with meanIoU as the performance metric. Known $c$ indicates whether the object center $c$ is known and center-aligned to the proposed region. Note that 0\%, 20-40\%, 40-60\%, and 60-80\% of the object are occluded in the respective FG Occlusion Levels and 1-20\%, 20-40\%, and 40-60\% of the context are occluded in the respective BG Occlusion Levels. Finally, PCNet-M is given additional ground truth occluder segmentation as supervision during inference, as noted by \textit{*}.}
\label{tab:amodal_occveh}

\end{table*}

\begin{table}[h]
\centering
\tabcolsep=0.088cm

\begin{tabular}{V{2.5}lV{2.5}cV{2.5}cV{2.5}cV{2.5}cV{2.5}cV{2.5}cV{2.5}cV{2.5}}
		\hline
		\multicolumn{8}{V{2.5}cV{2.5}}{Amodal Segmentation on KINS} \\
		\hline
		\multirow{2}{*}{Methods} &\multirow{2}{*}{k. $c$} &\multirow{2}{*}{superv.} &\multirow{2}{*}{FG-0} &\multirow{2}{*}{FG-1} &\multirow{2}{*}{FG-2} &\multirow{2}{*}{FG-3} &\multirow{2}{*}{Mean} \\
		& & & & & & & \\
		\hline  
		PCNet-M         &\xmark & \textit{mask} &\textbf{75.3}	&65.5	&52.9	&33.5	&56.8\\
		Ours-ML        	&\xmark & \textit{box} &69.2	&\textbf{68.7}	&62.7	&45.2	&61.5\\
		Ours-E2E        &\xmark	& \textit{box} &69.9	&68.1	&\textbf{63.2}	&\textbf{47.3}	&\textbf{62.1}\\
		\hline	
		\hline
		BBTP            &\cmark & \textit{box} &\textbf{77}	&68.3	&58.9	&53.9	&64.5\\
		Ours-ML        	&\cmark & \textit{box} &71.8	&\textbf{70.1}	&\textbf{66.2}	&57.8	&66.5\\
		Ours-E2E        &\cmark & \textit{box} &72.3	&69.6	&\textbf{66.2}	&\textbf{58.5}	&\textbf{66.7}\\
	    \hline
\end{tabular}

\caption{Amodal Segmentation performance is evaluated on the KINS dataset with meanIoU as the performance metric. ``k.$c$" indicates whether the object center $c$ is known during inference.  indicates whether the object center $c$ is known. Note that 0\%, 1-30\%, 30-60\%, and 60-90\% of the object are occluded in the respective Foreground Occlusion Levels.}
\label{tab:amodal_kins}
\end{table}

\subsection{Experimental Setup}
\textbf{Datasets.} 
Following the experimental settings of related work \cite{compnet_det_cvpr20}, we categorize the occluded objects in each dataset into three levels of foreground occlusion from FG-1 to FG-3 and, if applicable, into three levels of background occlusion from BG-1 to BG-3. 

The \textit{OccludedVehicles} dataset \cite{compnet_det_cvpr20} extends PASCAL3D+ \cite{xiang2014beyond} with synthetic occlusion. It contains 51801 objects evenly distributed among all occlusion levels, with both foreground and context occluded by unseen occluders. 

The \textit{KINS} dataset \cite{kins} contains real occlusion with amodal annotations. We restrict the scope of the evaluation to vehicles with a minimum height of 50 pixels, since the relevance of segmentation decreases as resolution reduces. Finally, the evaluation set contains 14826 objects.

The \textit{COCOA-cls} dataset \cite{cocoacls} is an extension of \textit{Amodal COCO} \cite{amodal_coco} with class annotations, totalling 766 objects.

The \textit{Occluded COCO} dataset \cite{compnet_cls_cvpr20} was introduced to test robustness of image classification to partial occlusion. It contains partially occluded objects from MS-COCO \cite{lin2014microsoft}.

\textbf{Baselines.}
As there is no existing model that performs amodal segmentation with class/box-level supervision only, we benchmark our model against \textit{BBTP} \cite{bbtp_hsu2019weakly}, a state-of-the-art weakly-supervised segmentation approach, and \textit{PCNet-M} \cite{deocclusion_2020_CVPR}, a self-supervised approach that leverages artificially generated amodal segmentation masks for training. 

\textit{BBTP} explores bounding box tightness prior to generate object mask under box-supervision and requires the input bounding box to be aligned to the object center $c$.

\textit{PCNet-M} utilizes Mask RCNN \cite{he2017mask} as an instance segmentation backbone and learns amodal completion by artificially occluding objects with other objects from the same dataset in a self-supervised manner. Hence, \textit{PCNet-M} is considered to be the mask-supervised upper bound for our model. 
Since both \textit{PCNet-M} and our model only leverage the visible parts of the object, they do not require known object center $c$. 

\textbf{Evaluation.}
It is observed that the occlusion levels in \textit{KINS} are severely disproportional: over $62\%$ of the objects are non-occluded and less than $8\%$ of objects are more than 60\% occluded. Therefore, in order to examine the mask prediction quality as a function of occlusion level, we evaluate with the best region proposals (highest IoU to ground truth) generated by an RPN as supervision, removing bias towards non-occluded objects in other metrics like mAP, and separate objects into subsets based on their occlusion level. 

Finally, due to the limited number of annotated objects in \textit{COCOA cls.}, we combine the train and test set and use the combined dataset to evaluate how well models can transfer to a novel domain when trained on \textit{OccludedVehicles}.

\begin{table}
\captionsetup{skip=0pt}
\centering
\tabcolsep=0.088cm
\begin{center}
\begin{tabular}{V{2.5}lV{2.5}cV{2.5}cV{2.5}cV{2.5}cV{2.5}cV{2.5}cV{2.5}cV{2.5}}
		\hline
		\multicolumn{8}{V{2.5}cV{2.5}}{Amodal Segmentation on COCOA \textit{cls.}} \\
		\hline
		\multirow{2}{*}{Methods} &\multirow{2}{*}{k. $c$} &\multirow{2}{*}{superv.} &\multirow{2}{*}{FG-0} &\multirow{2}{*}{FG-1} &\multirow{2}{*}{FG-2} &\multirow{2}{*}{FG-3} &\multirow{2}{*}{Mean} \\
		& & & & & & & \\
		\hline  
		PCNet-M         &\xmark & \textit{mask}  &56.8	&53.6	&47	&38.4	&49\\
		Ours-ML         &\xmark & \textit{box}  &\textbf{61.1}	&\textbf{62}	&\textbf{60}	&\textbf{54.3}	&\textbf{59.4}\\
		Ours-E2E        &\xmark & \textit{box}  &58.3	&59.8	&58.6	&53.5	&57.6\\
		\hline	
		\hline
		BBTP            &\cmark & \textit{box}  &57.3	&49.4	&40.7	&35	    &45.6\\
		Ours-ML         &\cmark & \textit{box}   &65	&64.2	&64.2	&60.9	&63.6\\
		Ours-E2E        &\cmark & \textit{box}  &\textbf{65.3}	&\textbf{65}	&\textbf{64.3}	&\textbf{61.4}	&\textbf{64}\\
	    \hline
\end{tabular}
\end{center}
\caption{Transfer Evaluation from \textit{OccludedVehicles} to \textit{COCOA cls.} with meanIoU as the performance metric. ``k.$c$" indicates whether the object center $c$ is known during inference.  indicates whether the object center $c$ is known during inference. Note that 0\%, 1-20\%, 20-40\%, and 40-70\% of the object are occluded in the respective Foreground Occlusion Levels.}
\label{tab:amodal_coco}
\end{table}

\begin{figure*}[t]
\centering
    \subfloat[\centering Known Object Center Comparison]{{\includegraphics[width=8.57cm]{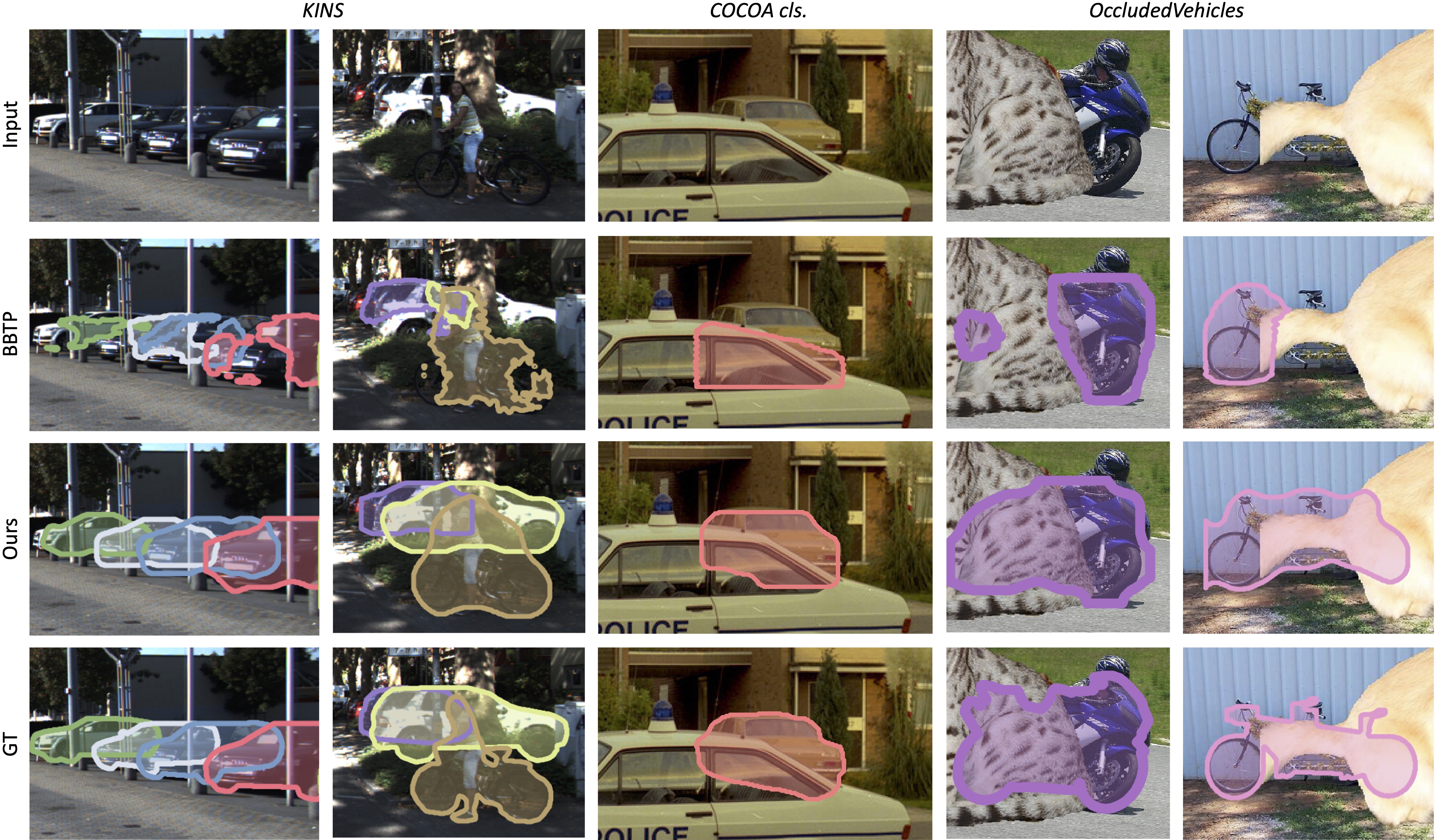} }} 
    $\mkern2.5mu$
    \subfloat[\centering Unknown Object Center Comparison]{{\includegraphics[width=8.57cm]{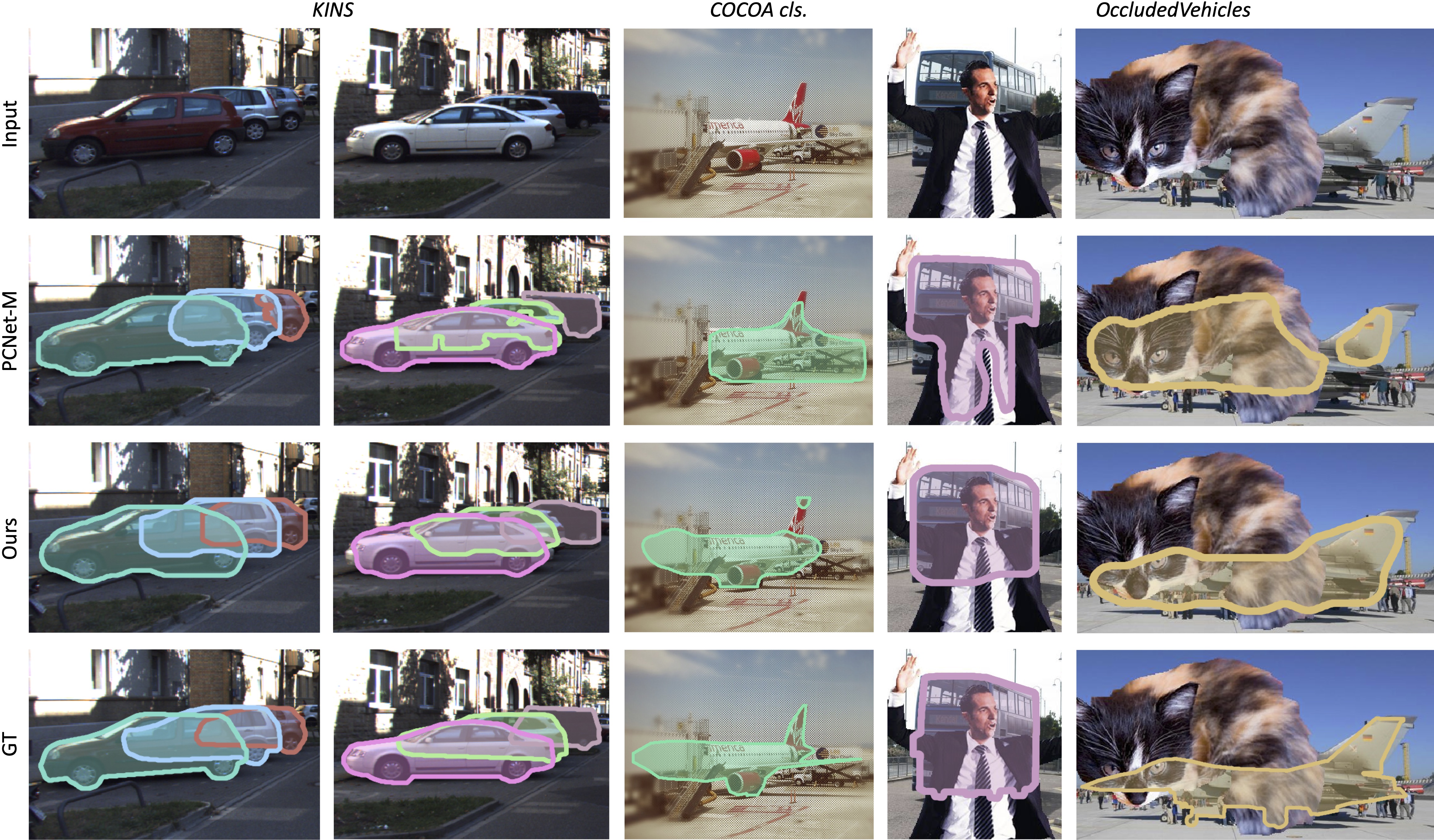} }}
    \caption{Qualitative Amodal Segmentation Results. For cases of known and unknown ground truth object centers, we present the raw image, BBTP/PCNet-M predictions, our model's predictions, and Ground Truth from the first to fourth row, respectively.}
    \label{fig:qualitative}
\end{figure*}

\textbf{Model and Training setup.} 
Since our Bayesian generative model is first learned with Maximum Likelihood and then fine-tuned in an end-to-end manner, we evaluate both separately, denoted as \textit{Ours-ML} and \textit{Ours-E2E} respectively.

\textit{Ours-ML}. Our model is initially learned from feature activations ($l = 4$) of a ResNeXt-50 \cite{xie2017aggregated} model, pretrained on ImageNet \cite{deng2009imagenet}. Specifically, we initialize compositional parameters $\{\mu_k\},\mathcal{A}, \mathcal{Z}, P(\vec w|y,m)$ and set the vMF variance to $\sigma_k = 65, \forall k \in \{1,\dots,K\}$, and the number of mixtures to $M=8$. We also learn the parameters of $n=5$ outlier models in an unsupervised manner and fixed a prior. During initialization, we optimize the parameters with an EM algorithm as described in Section \ref{sec:ourmodel}. 

\textit{Ours-E2E}. After learning via Maximum Likelihood, we use the obtained solution as initialization and fine-tune the model parameters as described in Section \ref{sec:e2e}. We choose AdaGrad \cite{duchi2011adaptive} with momentum $r=0.98$, a learning rate of $lr=0.01$, and trade-off weights $\gamma_1=2, \gamma_2=1$ for $10$ epochs on one NVIDIA TITAN Xp for a total of 2 hours.   

\subsection{Results with known object center}
\label{sec:amodal}

As \textit{BBTP} is assuming the full object bounding box (including the invisible part of the object), it is not able to estimate the amodal segmentation without knowing the object center and the corresponding full bounding box at test time. Thus, in order to evaluate and compare against \textit{BBTP}, the object center $c$ is given as supervision during inference, even though our model does not necessarily require object center $c$ to estimate the amodal segmentation. 

\textbf{Synthetic Occlusion.} As shown on the \textit{OccludedVehicles} dataset (Table \ref{tab:amodal_occveh}), both our Bayesian model learned via Maximum Likelihood (\textit{Ours-ML}) and fine-tuned via end-to-end training (\textit{Ours-E2E}) outperform \textit{BBTP} in amodal segmentation in all but two occlusion settings. Notably, in the highest occlusion level (FG-3 BG-3), our fine-tuned model is able to outperform \textit{BBTP} by more than 13\% in meanIoU.

\textbf{Real Occlusion.} Additionally, the trend observed in Table \ref{tab:amodal_occveh} can be confirmed by results under realistic occlusion. When evaluated on the \textit{KINS} dataset (Table \ref{tab:amodal_kins}), both of our models outperform \textit{BBTP} across all occlusion settings. Similarly, in the highest occlusion level, \textit{Ours-E2E} outperforms \textit{BBTP} by more than 4\% in meanIoU.

\textbf{Transferability.} Seen in Table \ref{tab:amodal_coco}, our models learned via Maximum Likelihood and fine-tuned end-to-end outperform \textit{BBTP} across all occlusion settings when learned from \textit{OccludedVehicles} and transferred to \textit{COCOA cls}. Notably, our end-to-end fine-tuned model outperforms \textit{BBTP} in domain generalization and surpasses it by more than 18\% in meanIoU on average. Furthermore, the observed increase in performance across all occlusion levels with known center when our model is only fine-tuned on unoccluded images further demonstrates the efficacy of the Maximum Likelihood loss term introduced in Section \ref{sec:e2e}.  

Qualitatively, shown in Figure \ref{fig:qualitative} (a), it is apparent that the mask proposals generated by \textit{BBTP} are negatively affected by the presence of occluders, while our proposed model can accurately estimate the object's amodal segmentation and preserve the object's shape consistency.

In conclusion, both quantitative and qualitative results with known object center demonstrate that our proposed model outperforms the state-of-the-art weakly-supervised method by a wide margin at amodal instance segmentation and out-of-domain transferability.

\subsection{Results with unknown object center}
\label{sec:modal}

In contrast to the previous section, since \textit{PCNet-M} is trained with annotated occlusion, both of our models, \textit{Ours-ML} and \textit{Ours-E2E}, and \textit{PCNet-M} are evaluated without the object center $c$ given as supervision.

\textbf{Synthetic Occlusion.} \textit{PCNet-M} can only perform amodal segmentation when the class label of the occluder is known a-priori in the dataset. Therefore, \textit{PCNet-M} is inherently unsuitable to be evaluated on the \textit{OccludedVehicles} dataset, as all occluders in the dataset belong to unseen/novel categories without explicit class annotations. Hence, in order to evaluate \textit{PCNet-M}, we provide the \textit{ground-truth occluder segmentation at inference time} (marked as \textit{mask*} supervised). In contrast, our approach does not require any additional information about the occluder. 
Seen in Table \ref{tab:amodal_occveh}, even with given ground truth occluder segmentation during inference, the mask-supervised \textit{PCNet-M} still performs worse compared to our weakly-supervised model in meanIoU at the highest occlusion level.

\textbf{Real Occlusion.} Furthermore, the results on the \textit{OccludedVehicles} dataset is verified in the \textit{KINS} dataset (Table \ref{tab:amodal_kins}), where both of our models outperform \textit{PCNet-M} across all occluded settings. In the highest occlusion level, \textit{Ours-E2E} outperforms \textit{PCNet-M} by more than 13\% in meanIoU.

\textbf{Transferability.} Similar to Section \ref{sec:amodal}, both of our models outperform \textit{PCNet-M} on \textit{COCOA-cls} across all occlusion settings when transferred from \textit{OccludedVehicles}. Notably, since the E2E model is fne-tuned with known object center and object center is unknown, our model learned from Maximum Likelihood generalizes better and surpasses \textit{PCNet-M} by more than 10\% in meanIoU on average.

Qualitatively, observed in Figure \ref{fig:qualitative} (b), the mask-supervised \textit{PCNet-M} failed to predict accurate the amodal mask of occluded objects, while our model estimates the amodal regions accurately by leveraging the prior distribution of the object's shapes in the mask predictions. Specifically in the right two columns of Figure \ref{fig:qualitative} (b), our model is able to predict a much more realistic amodal segmentation, even though \textit{PCNet-M} uses an additional information of a given ground-truth occluder segmentation during inference.

To conclude, our Bayesian approach outperforms the PCNet-M baseline at high occlusion levels while only requiring box-level supervision.

\begin{table}
\centering
\tabcolsep=0.09cm
\begin{center}
\begin{tabular}{V{2.5}lV{2.5}cV{2.5}cV{2.5}c|c|c|cV{2.5}cV{2.5}}
		\hline
		\multicolumn{8}{V{2.5}cV{2.5}}{Shape Priors Ablation} \\
		\hline
		Methods & k. $c$ &superv.	&FG-0 &FG-1 &FG-2 &FG-3 & Mean \\
		\hline  
		w/o priors   &\cmark &\textit{box}   &61.6	&59.5	&58.7	&58.3	&59.5	\\
		w/ priors    &\cmark &\textit{box}   &68.3	&66.6	&65.9	&65	&66.5	\\
		\hline	
		\hline	
		gt. priors   &\cmark &\textit{mask}   &71.6	&69.5	&68.7	&67.6	&69.4	\\
	    \hline
\end{tabular}
\end{center}

\caption{Shape priors ablation is evaluated on the OccludedVehicles via meanIoU. Note that we report the mean performance across all BG Occlusion levels for each FG Occlusion level.}
\label{tab:ablation}
\end{table}

\begin{table}
\centering
\tabcolsep=0.05cm
\begin{center}
\begin{tabular}{V{2.5}lV{2.5}cV{2.5}cV{2.5}c|c|c|cV{2.5}cV{2.5}}
		\hline
		\multicolumn{8}{V{2.5}cV{2.5}}{Classification on Occluded COCO } \\
		\hline	
		Methods & k.$c$ &superv.	&FG-0 &FG-1 &FG-2 &FG-3 & Mean \\
		\hline
		ResNeXt-50  &\cmark &\textit{box}    &\textbf{97.4}	&85.5	&81.9	&56.3	&80.3 \\
		CompNet  &\cmark &\textit{box}     &94.9	&89.6	&84.6	&\textbf{65.8}	&\textbf{83.7} \\
		CA-CompNet  &\cmark  &\textit{box}   	    &96	&88.4	&81.1	&64.4	&82.5 \\
		Ours-ML   &\cmark  &\textit{box}   	&95	&\textbf{90.4}	&84	&63	&83.1 \\
		Ours-E2E  &\cmark &\textit{box}      &94	&89.6	&\textbf{85}	&\textbf{65.8}	&83.6 \\
	    \hline
\end{tabular}
\end{center}
\caption{Classification performance evaluated on Occluded COCO. Note that 0\%, 20-40\%, 40-60\%, and 60-80\% of the object are occluded in the respective FG Occlusion Levels.}
\label{tab:cls_coco}
\end{table}

\subsection{Ablation}
\label{sec:ablation}
In Table \ref{tab:ablation}, we evaluate the effects of shape priors on amodal segmentation on the \textit{OccludedVehicles} dataset by (1) ablating the priors (\textit{w/o prior}), and by (2) learning the priors with ground truth segmentation (\textit{gt. prior}). Seen in Table \ref{tab:ablation}, amodal segmentation using shape priors learned from bounding box annotations significantly outperforms that without shape priors, and give comparable results as using priors learned from ground truth mask annotations.

\textbf{Image Classification.} Since our model uses supervision for object classification only and generalizes out-of-task to infer object segmentation, we verify the image classification performance of our model relative to related Bayesian generative models (\textit{CompNet} \cite{compnet_cls_cvpr20} and \textit{CA-CompNet} \cite{compnet_det_cvpr20}) and a DCNN classifier with the same backbone under the same supervision.
Seen in Table \ref{tab:cls_occveh}, our model outperforms the classifier in \textit{BBTP} by more than 9\% in classification accuracy, and outperforms the classifier in \textit{PCNet-M} by more than 3\% in classification accuracy when the object center $c$ is unknown. Moreover, our model performs on-par with CompNet and CA-CompNet. Similarly, found in Table \ref{tab:cls_coco}, our model outperforms ResNeXt-50 by more than 3 \% when evaluated on Occluded COCO with real occlusions. 

In summary, our model performs favorably over BBTP and PCNet-M in terms of image classification. It also performs on par with CompNets but can additionally perform amodal perception reliably, while being trained from bounding box and class-level supervision only.

\begin{table}
\centering
\tabcolsep=0.05cm
\begin{center}
\begin{tabular}{V{2.5}lV{2.5}cV{2.5}cV{2.5}c|c|c|cV{2.5}cV{2.5}}
		\hline
		\multicolumn{8}{V{2.5}cV{2.5}}{Classification on OccludedVehicles} \\
		\hline	
		Methods & k. $c$ &superv.	&FG-0 &FG-1 &FG-2 &FG-3 & Mean \\
		\hline
		PCNet-M      & \xmark & \textit{mask}  &\textbf{98.7}	&\textbf{95.9}	&86.1	&59.2	&85 \\
		CompNet      & \xmark & \textit{box} &97.7	&93.6	&\textbf{87.3}	&\textbf{73.6}	&\textbf{88.1}\\
		CA-CompNet   & \xmark & \textit{box} &97.7	&93.4	&87	&73.3	&87.9\\
		Ours-ML      & \xmark & \textit{box} &97.7	&93.4	&86.8	&72.6	&87.6\\
		Ours-E2E     & \xmark & \textit{box} &97.8	&93.4	&87.2	&73.5	&88\\
	    \hline
	    \hline 
	    BBTP        & \cmark & \textit{box}  &\textbf{99.1}	&\textbf{96.6}	&86	&53.9	&83.9\\
		CompNet     & \cmark & \textit{box}  &97.8	&94.9	&90.8	&79.6	&90.8\\
		CA-CompNet  & \cmark & \textit{box}  &98.3	&95	&89.7	&76.6	&89.9\\
		Ours-ML     & \cmark & \textit{box}  &97.8	&95.2	&90.7	&80.2	&91\\
		Ours-E2E    & \cmark & \textit{box}  &98.3	&95.6	&\textbf{91.4}	&\textbf{81.4}	&\textbf{91.7}\\
	    \hline
\end{tabular}
\end{center}
\caption{Classification performance evaluated on OccludedVehicles. Note that a mean is taken across all BG Occlusion levels.}
\label{tab:cls_occveh}
\end{table}

\section{Conclusion}

In this work, we studied the problem of amodal segmentation from the perspective of out-of-task and out-of-distribution generalization with a Bayesian model. 
We learn a Bayesian generative model of neural network features, which explicitly represents the object’s background context and foreground shape.
This enables the model to localize occluded object parts and predict the occluded object shape.
Our Bayesian approach for amodal segmentation only requires bounding box and class supervision, achieving state-of-the-art performance at amodal segmentation when compared to other weakly-supervised method and even outperforming fully supervised methods at high occlusion levels. 

\textbf{Limitations and Societal Impact.}
One limitation of our work is the dependence on 2D shape priors, which would require a large number to properly represent highly non-rigid objects such as humans or animals. For future work, we therefore expect that the learning of 3D shape priors would render the model more efficient and would also enhance the generalization ability to previously unseen 3D poses.
Like most segmentation works, our work does not introduce any foreseeable societal impacts, but will generally promote more data-efficient and robust computer vision models.

\textbf{Acknowledgements.} We gratefully acknowledge funding support from Office of Naval Research (N00014-21-1-2812) and National Science Foundation (BCS-1827427).



{\small
\bibliographystyle{ieee_fullname}
\bibliography{egbib}
}

\end{document}